\documentclass[fleqn,10pt]{wlscirep}
\usepackage[utf8]{inputenc}
\usepackage[T1]{fontenc}
\usepackage{graphicx}%
\usepackage{multirow}%
\usepackage{amsmath,amssymb,amsfonts}%
\usepackage{amsthm}%
\usepackage{mathrsfs}%
\usepackage[title]{appendix}%
\usepackage{xcolor}%
\usepackage{textcomp}%
\usepackage{manyfoot}%
\usepackage{booktabs}%
\usepackage{algorithm}%
\usepackage{algorithmicx}%
\usepackage{algpseudocode}%
\usepackage{listings}%
\theoremstyle{thmstyleone}%
\theoremstyle{thmstyletwo}%
\theoremstyle{thmstylethree}%
\usepackage{booktabs}
\usepackage{soul}
\definecolor{lightgrey}{RGB}{192, 192, 192}
\definecolor{lightred}{rgb}{1, 0.8, 0.8}
\definecolor{lightblue}{rgb}{0.8, 0.9, 1}
\newcommand*\colourcheck[1]{%
  \expandafter\newcommand\csname #1check\endcsname{\textcolor{#1}{\ding{52}}}%
}
\colourcheck{blue}
\colourcheck{green}
\colourcheck{red}
\usepackage{pifont}
\usepackage[most]{tcolorbox}
\usepackage{colortbl}

\usepackage{multicol}
\usepackage{makecell}

\usepackage{enumitem}
\usepackage{wrapfig}
\usepackage{titletoc}
\usepackage{lipsum}
\usepackage{fontawesome5}
\usepackage{mathtools}
\usepackage{float}
\usepackage{listings}
\usepackage{float}

\tcbset{
  aibox/.style={
    width=500pt,
    top=10pt,
    colback=white,
    colframe=black,
    colbacktitle=black,
    enhanced,
    center,
    attach boxed title to top left={yshift=-0.1in,xshift=0.15in},
    boxed title style={boxrule=0pt,colframe=white,},
  }
}

\makeatletter
\newtcolorbox[auto counter]{AIbox}[2][]{%
  aibox,
  title={Prompt~\thetcbcounter: #2},
  before upper={\protected@edef\@currentlabel{\thetcbcounter}},
  #1
}
\makeatother

\title{Supervising the search process produces reliable and generalizable information-seeking agents}

\author[1,*]{Guangzhi Xiong}
\author[2,*]{Qiao Jin}
\author[3]{Xiao Wang}
\author[2]{Yin Fang}
\author[1]{Haolin Liu}
\author[2]{Yifan Yang}
\author[4]{Fangyuan Chen}
\author[5]{Zhixing Song}
\author[6]{Dengyu Wang}
\author[3]{Minjia Zhang}
\author[2,+]{Zhiyong Lu}
\author[1,+]{Aidong Zhang}
\affil[1]{Department of Computer Science, University of Virginia, USA}
\affil[2]{National Library of Medicine, National Institutes of Health, USA}
\affil[3]{Department of Computer Science, University of Illinois Urbana–Champaign, USA}
\affil[4]{Medical Oncology, Dana–Farber Cancer Institute, USA}
\affil[5]{Surgery, University of Alabama at Birmingham, USA}
\affil[6]{Department of Neurology, Yale School of Medicine, USA}

\affil[*]{These authors contributed equally to this work.}

\affil[+]{Co-correspondence.}

\begin{abstract}    
Large language models (LLMs) are transforming web search by shifting from document ranking to synthesizing answers, and are increasingly deployed as autonomous agentic search systems that iteratively interact with external knowledge sources. Despite this progress, building effective search agents remains challenging because high-quality intermediate search steps are difficult to generate. Previous approaches have primarily relied on outcome supervision, rewarding agents only for producing correct final answers. This often leads to reward hacking and excessive dependence on parametric memory, limiting generalization to out-of-domain tasks. To address these limitations, we introduce RAG-Gym, a framework that shifts supervision from final answers to the search process itself. With RAG-Gym, we systematically investigate architecture design, parameter optimization, and action evaluation, identifying reasoning reflection as a critical capability for search agents. Building on this insight, we propose Re$^2$Search++, a process-supervised agent that achieves substantial improvements on multi-hop information-seeking benchmarks, especially in out-of-domain settings. Performance gains are driven primarily by higher-quality search queries rather than answer optimization alone, and the learned search critics transfer across models, including proprietary LLMs. These findings show that supervising the search process produces more reliable and generalizable information-seeking agents.
\end{abstract}

\begin{document}

\flushbottom
\maketitle
\thispagestyle{empty}

\section{Introduction}

Large language models (LLMs) are fundamentally transforming how humans seek information from the Internet. We are witnessing a paradigm shift from traditional search engines, which return lists of documents for users to synthesize, to generative answer engines that synthesize knowledge directly \cite{suri2024use,venkit2024search,li2024survey,hersh2024search,spatharioti2025effects}. This transition is largely powered by Retrieval-Augmented Generation (RAG), which grounds model outputs in external documents to ensure factuality \cite{lewis2020retrieval,gao2023retrieval}. Standard RAG systems typically operate in a single pass, retrieving once and generating immediately. This approach inherently limits their ability to solve complex, multi-hop problems that require gathering and linking multiple pieces of evidence. To overcome this limitation, the field is advancing toward autonomous information-seeking agents \cite{li2025search,shi2025iterative,yao2023react,asai2023self}, where LLMs iteratively interact with external knowledge sources to complete search tasks \cite{shinn2024reflexion,trivedi2023interleaving,jiang2023active,press2023measuring,coelho2025deepresearchgym}. This shift marks a move from simple retrieval to autonomous research, enabling systems to navigate complex information environments in a human-like manner \cite{nahid2025prism,jiang2025retrieve,agashe2025agent,agashe2025agent2,zhou2025autonomous}.

Despite recent advances, building effective autonomous search agents remains a significant challenge. The core obstacle lies in the complexity of decision making required to navigate information environments \cite{lin2025comprehensive,huang2025manusearch}. To resolve multi-hop queries, an agent must break down high-level problems into verifiable intermediate steps, formulate precise queries, and synthesize discrete pieces of evidence. The search space for these intermediate actions is both vast and sparse. Because multi-step retrieval involves a combinatorial explosion of possible action sequences, most paths lead to dead ends or irrelevant information \cite{xiong2021answering,lan2020query,mitra2022constraint,jiang2023path}. Without explicit guidance, agents struggle to learn effective search strategies and often fall into unproductive search loops or rely excessively on parametric knowledge \cite{song2025demystifying,kim2025reliability}.
Building on the success of reinforcement learning in enhancing LLMs for mathematical reasoning \cite{shao2024deepseekmath,guo2025deepseek,wang2025reinforcement}, current methods attempt to address this challenge through outcome supervision, rewarding agents based on the correctness of their final response \cite{song2025r1,jin2025search,chen2025research}. While this approach can align model outputs with ground truth answers, it is susceptible to reward hacking and may incorrectly assign credit for intermediate actions, especially when answers are derived from parametric memory rather than intermediate evidence \cite{wang2025beyond2,lightman2024lets,jin2025beneficial}. As a result, these systems may fail to learn effective search strategies and often struggle to generalize, with performance degrading on out-of-domain tasks where internal knowledge is insufficient (Fig. \ref{fig:prm_vs_orm}), which reduces their practical utility.

In this work, we introduce RAG-Gym, a framework that moves the training paradigm from outcome-based signals to process-level supervision. We formulate agentic search as a Markov Decision Process (MDP) in which search queries act as macro-actions. This approach enables fine-grained supervision of intermediate search steps rather than focusing solely on final outcomes. Agents can therefore learn coherent and verifiable search trajectories that logically support their conclusions. Using this framework, we systematically decompose and optimize search agents along three key dimensions: architecture design, parameter optimization, and action evaluation. This allows us to identify the specific design choices that drive optimal performance.

We identify reasoning reflection as a powerful but underexplored capability for search agents and introduce a new agent architecture called Re$^2$Search (Reason, Reflection, Search) that embodies this approach. At each step, Re$^2$Search reasons toward the final answer and explicitly reflects on its reasoning process to uncover knowledge gaps. By integrating Re$^2$Search with the optimal strategies discovered through RAG-Gym, we develop Re$^2$Search++, a process-supervised agent that achieves strong performance on multi-hop and medical information-seeking benchmarks. Re$^2$Search++ surpasses robust outcome-supervised baselines \cite{song2025r1,jin2025search}, especially in out-of-domain evaluations.

Our analysis shows that these improvements arise from a greater ability to generate effective search queries, not just from optimizing answers. This is supported by substantial increases in the retrieval recall of ground-truth documents. We also find that the search heuristics learned through process supervision transfer well, enabling critics trained on open-source trajectories to guide proprietary models effectively. Supervising the evidence-gathering process, rather than focusing only on outcomes, provides a principled path to building reliable and generalizable information-seeking agents.

\section{Results} 

\subsection{A framework for supervising the search process}

\begin{figure}
    \centering
    \includegraphics[width=1\linewidth]{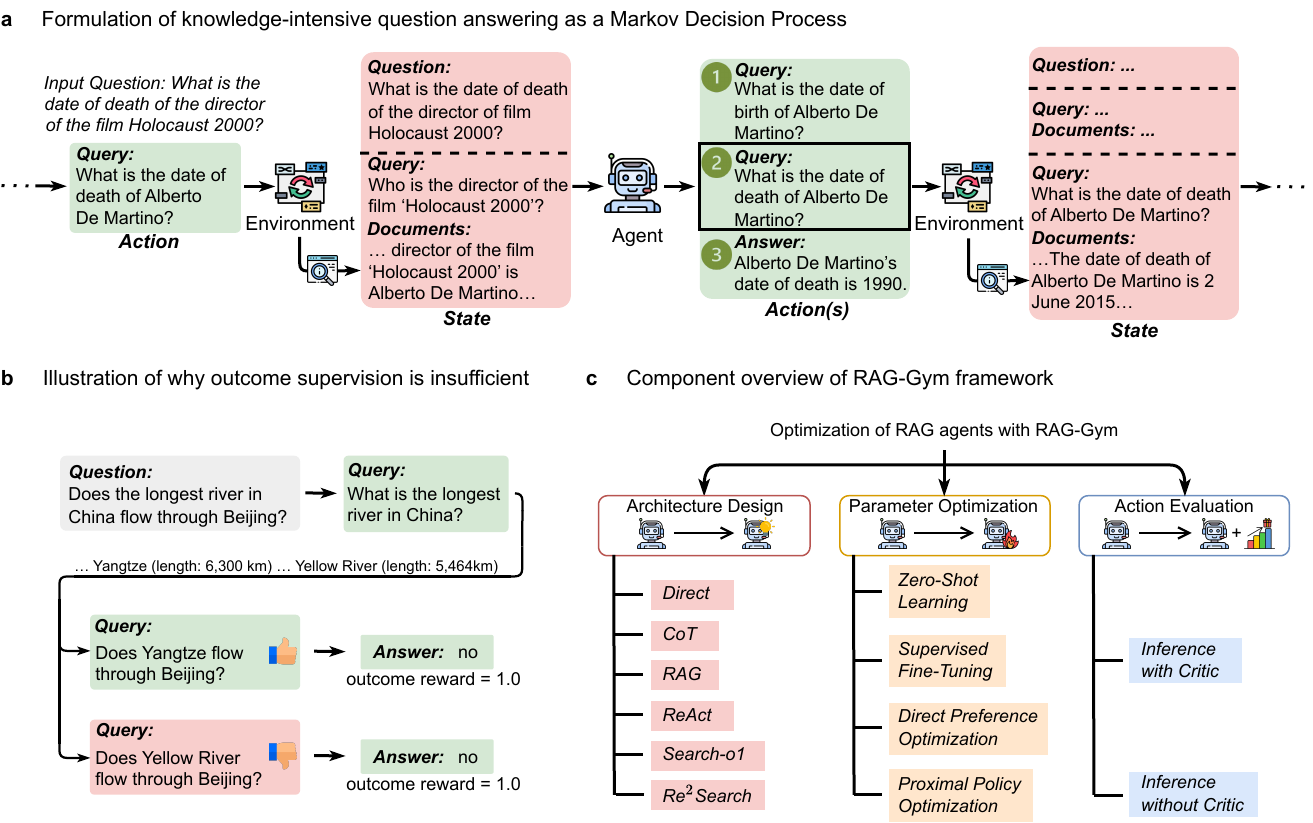}
    \caption{
    \textbf{Process supervision of search agents with RAG-Gym.}
    \textbf{a,} RAG-Gym formulates agentic search as a high-level Markov decision process (MDP). At each step $t$, the agent (Actor) receives a state $s_t$ comprising the question and history, and executes a macro-action $a_t$ (either issuing a search query or generating an answer), receiving an updated state $s_{t+1}$.
    \textbf{b,} Illustration of the limitation of outcome supervision. In outcome-only settings, a correct final answer assigns a positive reward to the entire trajectory, failing to penalize suboptimal intermediate steps or distinguish between unsupported answers and rigorously justified deductions.
    \textbf{c,} The RAG-Gym framework enables modular optimization across three dimensions: Architecture Design, Parameter Optimization, and Action Evaluation.
    }
    \label{fig:Fig1}
\end{figure}

We conceptualize agentic search as a high-level Markov Decision Process (MDP) rather than a standard token-by-token generation task. As illustrated in Fig. \ref{fig:Fig1}a, the agent functions as a policy interacting with an information-retrieval environment. At each step $t$, the state $s_t$ comprises the original question, the accumulated reasoning history, and the set of documents retrieved up to that point. The agent selects a macro-action $a_t$ by either issuing a targeted search query to gather additional evidence or terminating the process by generating a final answer.

This MDP formulation is crucial because it exposes intermediate reasoning steps to direct optimization. In contrast, standard approaches rely on outcome-only supervision (Fig. \ref{fig:Fig1}b), which suffers from a severe credit assignment problem. A correct final answer may result from unsupported or hallucinated knowledge and can incorrectly reinforce a flawed search trajectory. Conversely, a rigorous and necessary search step might be unfairly penalized if the subsequent answer generation is unsuccessful.

Building on the MDP formulation, we introduce RAG-Gym, a framework that decomposes the agentic design space into three coupled dimensions that enable process-level supervision (Fig. \ref{fig:Fig1}c). The first dimension is architecture design, which defines the agent's internal cognitive flow and verification loops. The second is parameter optimization, which trains the actor using supervised, preference, or reinforcement learning. The third is action evaluation, which trains an external critic to guide decision-making at inference. This modularity allows us to isolate the contributions of each component and move beyond the limitations of outcome-supervised baselines.

We evaluate our approach on four knowledge-intensive benchmarks: HotpotQA \cite{yang2018hotpotqa}, 2WikiMultihopQA \cite{ho2020constructing}, Bamboogle \cite{press2023measuring}, and MedQA \cite{jin2021disease}. For the first three, we report Exact Match (EM) and F1, and for MedQA, we report accuracy. Macro averages are computed by treating accuracy as EM and F1 in the multi-choice setting. Unless otherwise noted, process-reward data for actor and critic training consists of 1,000 questions sampled from HotpotQA or MedQA. Generalization is assessed on 2WikiMultihopQA and Bamboogle using models tuned on HotpotQA. Details about model training and evaluation are provided in the Methods.

\subsection{Reasoning-reflection improves search architecture}

\begin{figure}[h!]
    \centering
    \includegraphics[width=1.0\linewidth]{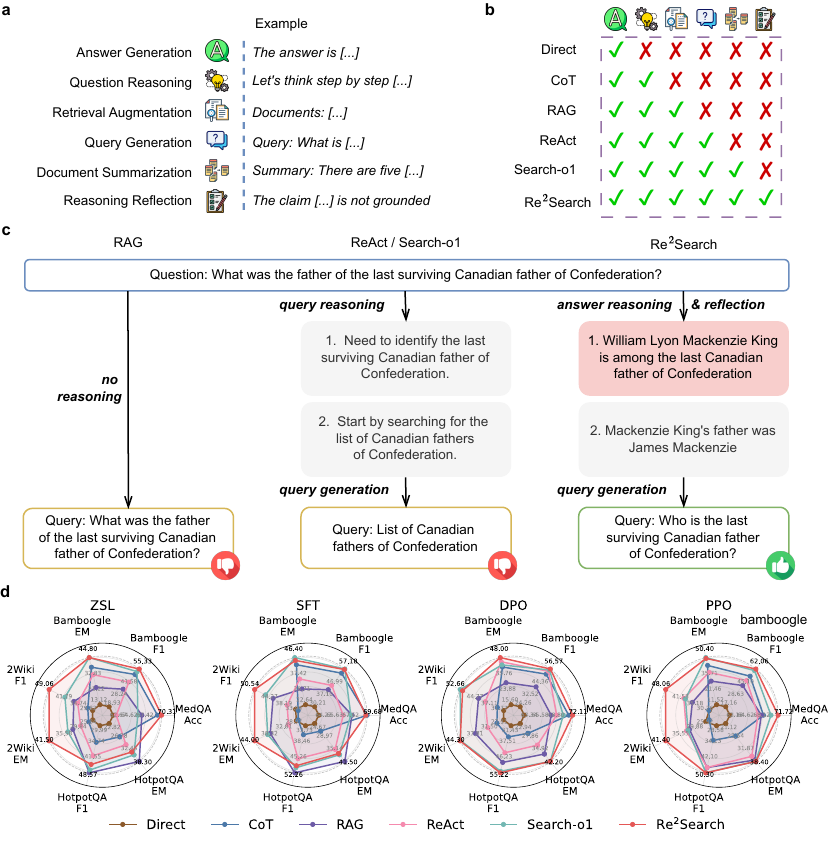}
    \caption{
    \textbf{Architecture components and optimization of information-seeking agents.}
    \textbf{a,} Decomposition of agent capabilities into six functional modules: answer generation, question reasoning, retrieval augmentation, query generation, document summarization, and reasoning reflection.
    \textbf{b,} Capability matrix contrasting baseline architectures (Direct, CoT, RAG, ReAct, Search-o1) against Re$^2$Search. Only Re$^2$Search instantiates the full functional stack, uniquely integrating the reasoning reflection module to verify intermediate steps.
    \textbf{c,} Illustrative trajectory for a complex multi-hop question. While standard agents proceed with generic queries, Re$^2$Search alternates between reasoning, explicit reflection on unverified claims (highlighted in red), and targeted follow-up queries.
    \textbf{d,} Performance radar charts across four benchmarks evaluating agents under ZSL and three post-training regimes (SFT, DPO, PPO). The Re$^2$Search architecture (red line) consistently encompasses the largest area, demonstrating robust superiority across datasets and optimization methods.
    }
    \label{fig:agent_architecture}
\end{figure}

To systematically address the limitations of current agentic search systems, we first decompose the cognitive requirements of search agents into six distinct functional modules: Answer Generation, Question Reasoning, Retrieval Augmentation, Query Generation, Document Summarization, and Reasoning Reflection (Fig. \ref{fig:agent_architecture}a). Existing architectures do not instantiate this full stack. For example, standard RAG systems retrieve based solely on keywords without reasoning, while existing search agents such as ReAct interleave thought and action but lack an explicit mechanism to target generated queries toward answer reasoning. As shown in Fig. \ref{fig:agent_architecture}b, only our proposed Re$^2$Search architecture incorporates the Reasoning Reflection module, which serves as a dedicated verification loop.

This mechanism fundamentally alters the search trajectory. Instead of immediately issuing queries based on topical associations, Re$^2$Search drafts an answer reasoning chain, inspects it for unverified claims, and transforms the first identified knowledge gap into a targeted query. The practical impact of this design is illustrated in Fig. \ref{fig:agent_architecture}c, where the agent attempts to identify the father of the last surviving Canadian father of Confederation. While standard agents such as ReAct and Search-o1 issue generic queries about the ``list of fathers,'' Re$^2$Search explicitly reasons that it must first identify the specific individual (e.g., William Lyon Mackenzie King is claimed but not verified in the original answer reasoning) before verifying his parentage. This ensures that every search action is grounded in logical necessity rather than heuristic guessing.

We validate this architectural advantage quantitatively across four knowledge-intensive benchmarks. The performance radar charts in Fig. \ref{fig:agent_architecture}d show that Re$^2$Search, represented by the red contour, consistently achieves the highest performance compared to five baselines. This dominance is observed across all training regimes, from Zero-Shot Learning (ZSL) to Proximal Policy Optimization (PPO), indicating that the Reasoning Reflection module provides a robust inductive bias that persists even after extensive parameter optimization.

\subsection{Process-level preference optimization enhances agent behavior}

\begin{figure}[h!]
    \centering
    \includegraphics[width=1.0\linewidth]{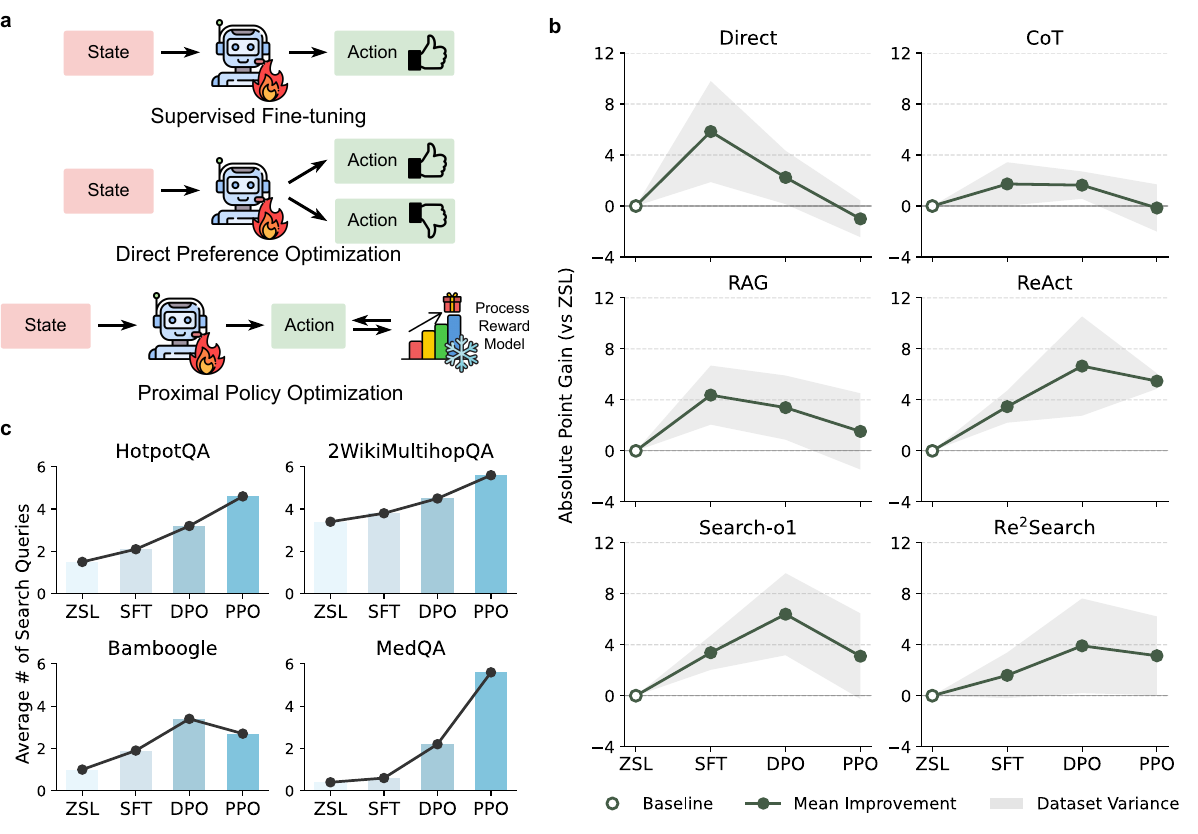}
    \caption{
    \textbf{Process supervision enhances multi-step reasoning and information-seeking behavior.}
    \textbf{a,} Illustration of three post-training objectives. Supervised fine-tuning (SFT) clones a reference trajectory. Direct preference optimization (DPO) contrasts preferred and less-preferred actions. Proximal policy optimization (PPO) updates the policy using a process-reward model trained on preference data.
    \textbf{b,} Performance distributions (F1 or Accuracy) across four datasets (HotpotQA, 2WikiMultihopQA, Bamboogle, MedQA). While simple agents (Direct, CoT, RAG) saturate with SFT, complex multi-step agents (ReAct, Search-o1, Re$^2$Search) achieve superior performance with DPO and PPO, highlighting the necessity of negative feedback in learning complex search strategies.
    \textbf{c,} Average number of search queries generated by Re$^2$Search agents during inference. Process-supervised agents consistently exhibit more active information-seeking behavior compared to zero-shot baselines (ZSL), correlating with improved answer quality.
    }
    \label{fig:supervision_algorithm}
\end{figure}

Having established the architectural advantages of reasoning-reflection, we next investigate how different supervision signals influence the agent's ability to navigate the information retrieval state space. We compare Zero-Shot Learning (ZSL) with three optimization regimes: Supervised Fine-Tuning (SFT), Direct Preference Optimization (DPO), and Proximal Policy Optimization (PPO), as illustrated in Fig. \ref{fig:supervision_algorithm}a.

Our analysis reveals a clear distinction in how supervision affects agent behavior, depending on the complexity of the agent's cognitive architecture. As shown in Fig. \ref{fig:supervision_algorithm}b, process supervision consistently improves performance over zero-shot baselines. However, the optimal training algorithm depends on the agent type. For single-step agents such as Direct, CoT, and standard RAG, SFT is sufficient to reach peak performance, and DPO or PPO provide little additional benefit. In contrast, agents capable of multi-step iterative search, including ReAct, Search-o1, and our Re$^2$Search, benefit substantially from preference or reinforcement learning. DPO in particular yields the largest improvements for Re$^2$Search, increasing F1 scores by an average of 3.92\% over baselines.

This difference suggests that for complex, multi-hop trajectories, simply imitating a correct path with SFT is not enough. The information-seeking search space contains far more unproductive actions, such as vague queries, redundant steps, or premature conclusions, than productive ones. As a result, the model benefits from the negative reward signals in DPO and PPO, learning explicitly which search actions to avoid. This performance gain is accompanied by a distinct behavioral shift. Agents trained with process supervision generate a higher volume of targeted queries during inference compared to their zero-shot counterparts, as shown in Fig. \ref{fig:supervision_algorithm}c. This increased activity indicates that the agent has learned to recognize its own knowledge boundaries and replaces hallucination with rigorous, step-wise verification.

\subsection{Critic-guided inference and cross-model transfer} \label{sec:critic}

\begin{figure}[h!]
    \centering
    \includegraphics[width=1.0\linewidth]{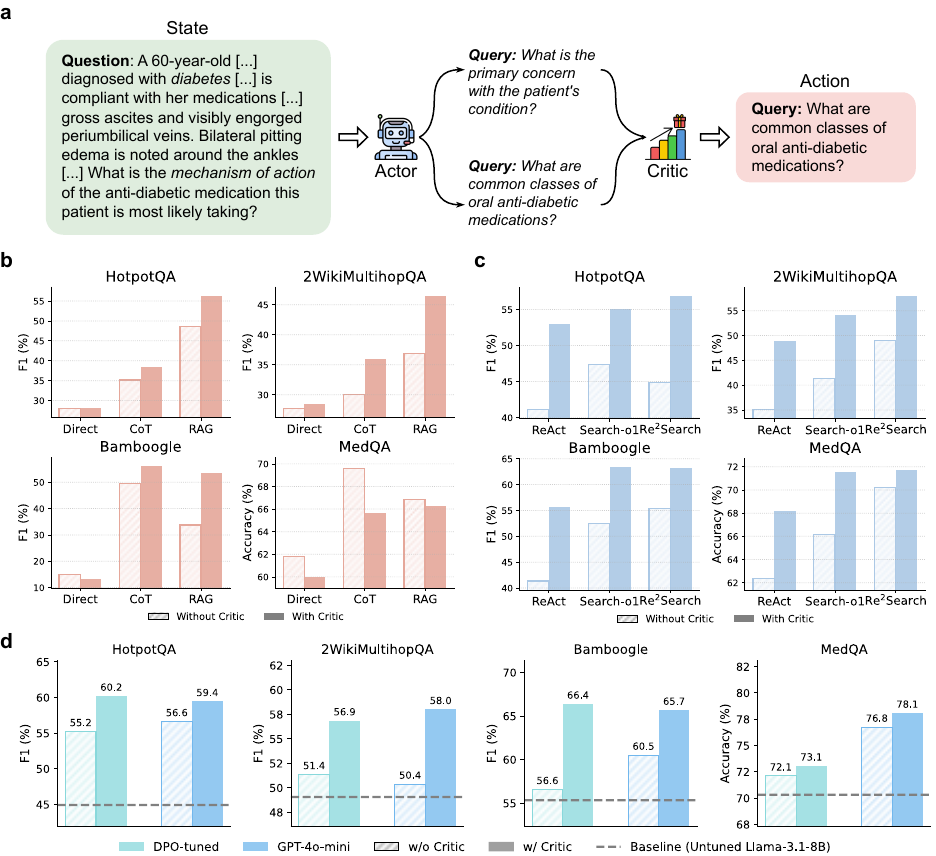}
    \caption{
    \textbf{Critic-guided action selection improves accuracy and transfers across model backbones.}
    \textbf{a,} Schematic of critic-guided inference in RAG-Gym. The actor proposes candidate actions (queries or answers), and a critic scores them to select the best action for the next step.
    \textbf{b,} Impact of critic-guided inference on single-step agents (Direct, CoT, RAG). While CoT and RAG see improvements on general-domain tasks (HotpotQA, 2WikiMultihopQA, Bamboogle), they suffer performance degradation on the domain-specific MedQA benchmark. The Direct prompting baseline shows negligible or negative impact across all tasks.
    \textbf{c,} Impact of critic-guided inference on multi-step agents (ReAct, Search-o1, Re$^2$Search). Unlike single-step agents, multi-step architectures achieve consistent performance gains across both general and medical domains.
    \textbf{d,} Cross-model transferability. A critic trained on Llama-3.1-8B trajectories improves both a DPO-tuned Llama-3.1-8B actor and a GPT-4o-mini actor, indicating that learned principles of ``good search'' generalize across model families without retraining the critic.
    }
    \label{fig:critic}
\end{figure}

To further refine the search process at runtime, we employ critic-guided action selection. Here, a reward model scores $N$ candidate actions at each step, and the agent executes the highest-scoring option using a Best-of-$N$ strategy. As illustrated in Fig. \ref{fig:critic}a, this mechanism steers the agent away from unproductive retrieval loops and toward verifiable evidence.

Our results show that the effectiveness of inference-time guidance depends on the structural complexity of the agent. On general-domain benchmarks, critic guidance improves F1 scores for reasoning-enhanced architectures such as CoT, RAG, ReAct, Search-o1, and Re$^2$Search, as shown in Fig. \ref{fig:critic}b and \ref{fig:critic}c. In contrast, simple Direct prompting does not benefit from this intervention. The limitation of single-step methods is even more pronounced on the domain-specific MedQA benchmark (Fig. \ref{fig:critic}b). In this setting, only multi-step agents such as ReAct, Search-o1, and Re$^2$Search achieve substantial gains, while single-step methods see limited improvements or even performance degradation. This distinction confirms that the critic primarily serves as a navigational aid, helping the agent traverse complex, sequential retrieval paths rather than merely acting as a static fact-checker for final answers.

We also find that the search heuristics learned by the critic are highly transferable across model backbones. In Fig. \ref{fig:critic}d, we deploy a critic trained solely on trajectories from Llama-3.1-8B to guide both a DPO-tuned Llama-3.1-8B actor and a distinct GPT-4o-mini actor. The Llama-trained critic improves the performance of the proprietary GPT-4o-mini model, for example increasing MedQA accuracy from 76.8\% to 78.1\%. This result indicates that the principles of effective information seeking are generalizable logic patterns rather than model-specific artifacts. Such transferability suggests a cost-effective deployment paradigm in which lightweight, open-source critics can govern the reasoning processes of larger, black-box systems without requiring access to their internal weights.

\subsection{Process supervision generalizes better than outcome supervision}

\begin{figure}[h!]
    \centering
    \includegraphics[width=1.0\linewidth]{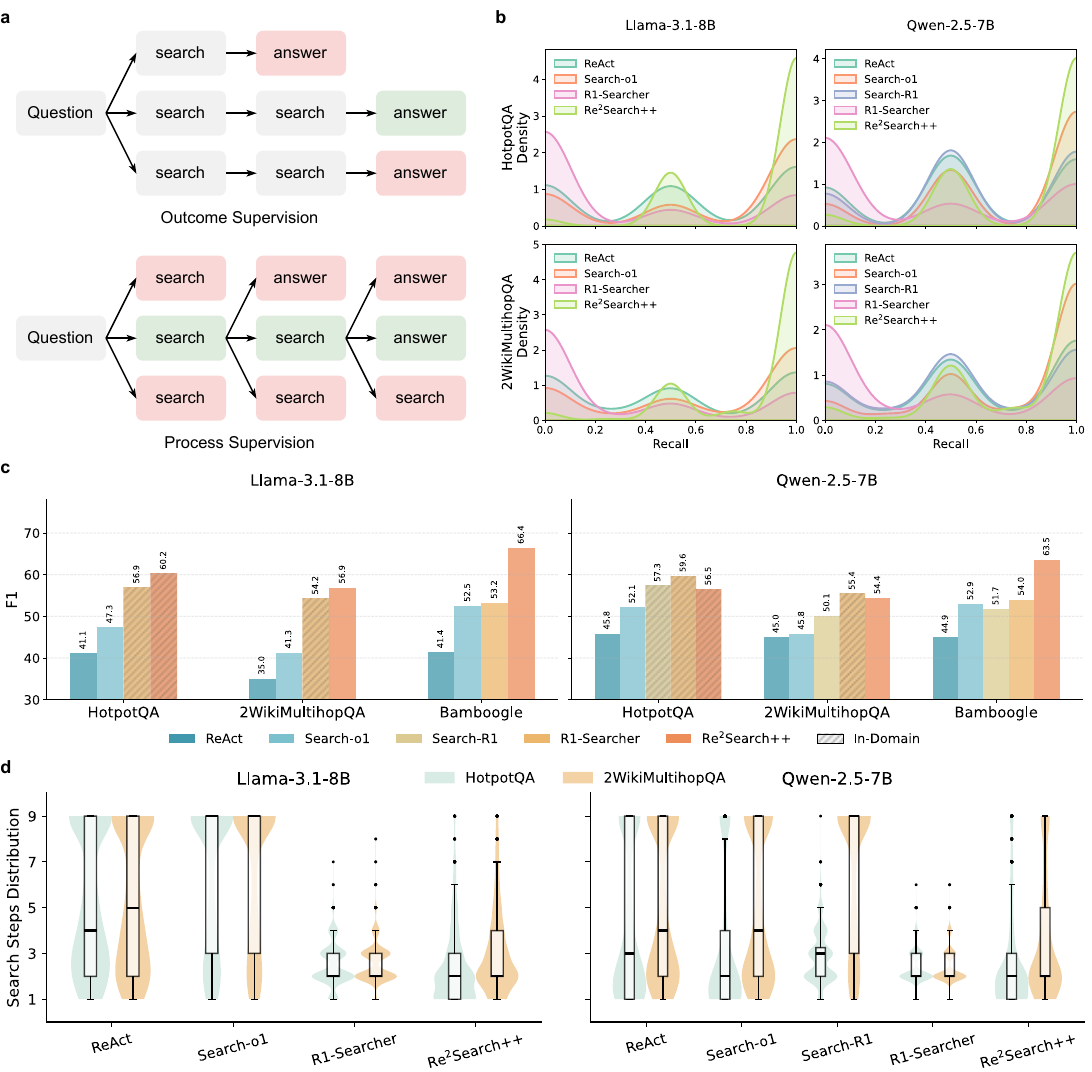}
    \caption{
    \textbf{Process supervision yields superior generalization and search efficiency compared to outcome supervision.}
    \textbf{a,} Conceptual contrast between outcome-only supervision, which rewards only the final answer, and process supervision, which assigns feedback at each intermediate search step.
    \textbf{b,} Retrieval recall distributions on HotpotQA and 2WikiMultihopQA for outcome-supervised baselines (ReAct, Search-o1, Search-R1, R1-Searcher) versus process-supervised Re$^2$Search++ (Llama-3.1-8B and Qwen-2.5-7B backbones). Process-supervised agents consistently achieve higher recall of ground-truth documents.
    \textbf{c,} Performance (F1) across in-domain and out-of-domain benchmarks. Re$^2$Search++ matches or outperforms outcome-supervised agents, including baselines tuned directly on the target dataset (``In-Domain''), and generalizes well to unseen benchmarks.
    \textbf{d,} Distribution of search steps per question. Re$^2$Search++ attains higher accuracy with comparable or fewer search actions, indicating a more targeted and efficient query strategy.
    }
    \label{fig:prm_vs_orm}
\end{figure}

We integrate the optimal components identified in our framework, namely the $Re^{2}$Search architecture, DPO training, and critic-guided inference, into a unified system called Re$^2$Search++. To isolate the impact of supervision style, we compare this process-supervised approach with strong outcome-supervised baselines, specifically Search-R1 and R1-Searcher, which reinforce trajectories based solely on the correctness of the final answer as shown in Fig. \ref{fig:prm_vs_orm}a.

Outcome supervision is effective for in-domain tasks such as HotpotQA, on which Search-R1 was explicitly trained. However, Fig. \ref{fig:prm_vs_orm}c demonstrates its limitations in generalization. Outcome-supervised baselines experience performance degradation on out-of-domain datasets. In contrast, Re$^2$Search++ achieves superior generalization, particularly on the unseen Bamboogle benchmark, while maintaining comparable or better performance than outcome-supervised agents on their respective in-domain tasks. This performance gap highlights the phenomenon of reward hacking that arises with outcome supervision. When supervision is sparse and provided only at the end of the trajectory, agents may learn to exploit parametric memory to guess answers and perform superficial searches that merely satisfy the training loop.

We further validate this hypothesis by analyzing retrieval recall distributions in Fig. \ref{fig:prm_vs_orm}b. Outcome-supervised agents frequently achieve low retrieval recall, indicating that their correct answers often result from ungrounded predictions or reliance on parametric memory rather than retrieved evidence. In contrast, Re$^2$Search++ demonstrates significantly higher retrieval recall across both Llama-3.1 and Qwen-2.5 backbones, showing that process supervision aligns the agent with the actual utility of retrieved evidence. As shown in Fig. \ref{fig:prm_vs_orm}d, this improved grounding does not come at the cost of efficiency. Re$^2$Search++ attains higher accuracy with comparable or fewer search actions than baselines, reflecting a more targeted query strategy. By rewarding the steps of evidence gathering, we produce an agent that is verifiable and faithful and remains effective even when encountering unfamiliar questions where internal knowledge is insufficient.

\subsection{Quality of process supervision signals}

The effectiveness of process supervision relies fundamentally on the quality of the reward signal used to train the critic. To explore this, we assessed four types of signals as shown in Fig. \ref{fig:discussion}a and \ref{fig:discussion}b. These include a baseline outcome-only reward model (ORM) and three process reward models (PRMs) based on random sampling, Monte Carlo rollouts \cite{wang2024math}, and model-based annotation using Llama-3.1-8B and GPT-4o.

Our findings indicate that the utility of a reward signal in retrieval-augmented generation (RAG) differs markedly from its behavior in pure reasoning tasks such as mathematics. Although rollout-based verification is widely used for logical reasoning, it performs poorly in the RAG context. For example, as shown in Fig. \ref{fig:discussion}d, the rollout-based PRM achieves an F1 score of 49.6\% on the 2WikiMultihopQA benchmark, which is lower than the outcome-only baseline at 55.6\% F1. This drop in performance can be traced to the ``false positive'' issue that arises in retrieval settings. An agent may produce a factually correct final answer by relying on parametric memory or exploiting spurious correlations, even if its search trajectory is irrelevant or hallucinated. Rollout methods, which validate based solely on the final answer, tend to reinforce these flawed trajectories without discrimination.

In contrast, model-based supervision, particularly when distilled from GPT-4o, demonstrates greater robustness to noise. As detailed in Fig. \ref{fig:discussion}c, GPT-4o annotations show the highest agreement with human ground-truth judgments, reaching 85.8\% on MedQA, and achieve high recall in retrieving ground-truth documents, such as 84.5\% on 2WikiMultihopQA. Critics trained on these signals consistently deliver the best downstream performance, achieving 57.9\% F1 on 2WikiMultihopQA and 72.0\% accuracy on MedQA. These results suggest that strong model-based annotators are able to distinguish between strictly justified answer trajectories and unjustified guesses, a distinction that outcome-dependent rollout methods do not capture.

\subsection{Scaling properties of data efficiency and inference compute}

To guide efficient deployment, we examine how RAG-Gym performance changes with training data volume and inference-time compute.

\begin{figure}[h!]
    \centering
    \includegraphics[width=0.95\linewidth]{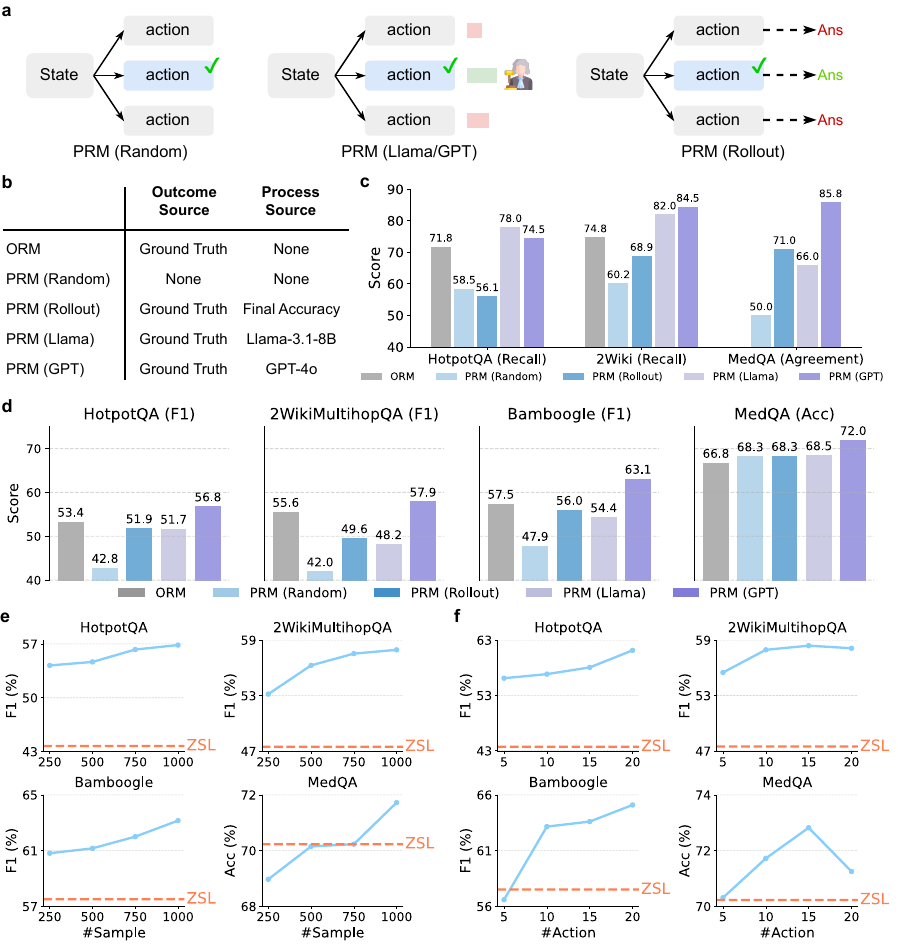}
    \caption{
    \textbf{Impact of reward signal quality and scaling properties of process supervision.}
    \textbf{a,} Illustration of different Process Reward Model (PRM) sources: Random (baseline), Model-based (Llama-3.1-8B/GPT-4o), and Rollout (Monte Carlo estimation).
    \textbf{b,} Categorization of reward sources based on ground truth access and supervision type.
    \textbf{c,} Evaluation of process quality. PRMs trained with GPT-4o annotations exhibit the highest agreement with ground truth and human judgment, whereas rollout methods underperform in RAG settings.
    \textbf{d,} Downstream performance (F1/Accuracy) using critics trained on different reward sources. GPT-4o-derived rewards consistently yield the best performing agents.
    \textbf{e,} Data efficiency scaling. General-domain tasks (HotpotQA, 2WikiMultihopQA, Bamboogle) show performance gains over the ZSL baseline (dashed line) with as few as 250 samples, while the medical-domain task (MedQA) requires larger datasets to converge.
    \textbf{f,} Inference compute scaling via Best-of-$N$ sampling. Performance typically improves with candidate pool size $N$, saturating between 15 and 20 actions.
    }
    \label{fig:discussion}
\end{figure}

\paragraph{Data Efficiency.}
We evaluate the critic's sample complexity by varying the number of process-rewarded training samples from 250 to 1,000. The learning dynamics differ between general and domain-specific tasks. On general-domain benchmarks such as HotpotQA, 2WikiMultihopQA, and Bamboogle, the critic quickly acquires effective search heuristics. Substantial F1 improvements over the Zero-Shot Learning (ZSL) baseline are achieved with as few as 250 samples. In contrast, domain-specific tasks like MedQA display a U-shaped learning curve, as shown in Fig. \ref{fig:discussion}e. Performance at 250 samples initially falls below the baseline, then recovers and peaks at 1,000 samples. This pattern suggests that while general information-seeking logic is rapidly learnable, recognizing valid domain-specific reasoning requires more extensive data.

\paragraph{Inference Compute.}
We also investigate the relationship between computational budget and performance using a Best-of-$N$ inference strategy. As illustrated in Fig. \ref{fig:discussion}f, increasing the candidate pool size $N$ leads to consistent performance gains up to a saturation point, typically between $N=15$ and $N=20$. Beyond this range, returns diminish and, in some cases such as MedQA, performance declines. For example, accuracy drops from 72.8\% at $N=15$ to 71.2\% at $N=20$. This decline likely results from a reward hacking effect, where a larger sample size increases the chance of generating semantically flawed but high-scoring adversarial actions that can mislead the critic. These results indicate that a moderate inference budget, with $N$ around 10 to 15, offers the best balance between accuracy and computational cost.

\section{Discussion}

Current agentic systems are typically optimized for the correctness of their final answers, often overlooking the reasoning process that leads to those answers. Our findings show that emphasizing process-level supervision, rather than focusing solely on outcomes, provides a more robust foundation for developing reliable information-seeking agents. By framing retrieval-augmented generation as a Markov decision process in which intermediate actions are both observable and evaluable, RAG-Gym enables direct optimization of the search process. This methodology consistently delivers performance improvements over strong baselines on multi-hop and domain-specific benchmarks. Our results suggest that supervising the search process of an agent is more effective for ensuring reliability than supervising only its final outputs.

A key insight from our study is the fragility of outcome-based reward models. Systems trained only on final correctness often exhibit reward hacking, where agents exploit parametric memory to guess answers while conducting only superficial searches to satisfy training objectives. This can result in agents that appear effective in familiar domains but fail to generalize when their internal knowledge is insufficient. In contrast, our process-supervised approach addresses this issue by encouraging a cycle of reasoning and reflection. By explicitly rewarding the identification of knowledge gaps and the formulation of targeted queries, the model learns to prioritize evidence gathering over shortcuts. Our recall analysis further supports this alignment, showing that process-supervised agents retrieve more relevant documents and that performance improvements are driven by genuine search utility rather than hallucinated reasoning.

The modular design of our framework offers a practical approach for deploying these systems by leveraging portable critics. We find that a critic trained on process data from a smaller open-source model can successfully guide the inference of entirely different backbone models. This demonstrates that the core principles of effective information seeking are generalizable logic patterns rather than model-specific artifacts. Such transferability suggests a cost-effective deployment paradigm in which lightweight, open-source critics can guide the reasoning processes of larger, black-box systems without requiring access to their internal weights.

While our results show that process supervision improves reliability and generalization, several challenges remain. The approach depends on high-quality reward signals, and we found that rollout-based rewards can be noisy in retrieval settings because coincidental answer correctness may create false positives. As a result, stronger model-based annotators are currently more effective, although they may introduce additional bias and cost. Our evaluation also focuses primarily on benchmark question-answering settings, and broader validation across open-ended search tasks, domains, and retrieval environments will be important in future work. In addition, critic-guided Best-of-N inference increases computational cost, although moderate candidate pool sizes provide a favorable trade-off between accuracy and efficiency. Extending this framework to broader agentic tasks such as tool use and planning remains an important direction for future research.

Overall, our results indicate that supervising the search process, rather than only the final answer, is a promising path toward building information-seeking agents that are more reliable, better grounded in retrieved evidence, and more likely to generalize beyond their training domains. Future work should explore stronger and more scalable sources of process supervision, more efficient inference-time guidance, and extensions from question answering to broader information-seeking and decision-making tasks.

\section{Materials and Methods}

We evaluate a family of search agents within a unified retrieval-augmented generation (RAG) environment, RAG-Gym. Agents differ along three axes: (i) architecture design, which determines which reasoning and retrieval modules are instantiated; (ii) post-training objective, which specifies how process-level feedback is used to tune the actor; and (iii) inference-time guidance, provided by a lightweight critic trained on process-reward data.

\subsection{Benchmark datasets and evaluation protocol}

We benchmark agents on four tasks that require extensive knowledge and reasoning. These include three multi-hop question answering datasets grounded in Wikipedia (HotpotQA \cite{yang2018hotpotqa}, 2WikiMultihopQA \cite{ho2020constructing}, and Bamboogle \cite{press2023measuring}) as well as a multiple-choice medical dataset, MedQA \cite{jin2021disease}. To ensure consistent comparisons and enable process-supervision training, we adopt standardized dataset splits.

For HotpotQA, we use the last 1,000 validation questions for evaluation and the first 1,000 for constructing the process-reward dataset that trains actors and critics. 2WikiMultihopQA contains multi-hop questions with systematically curated reasoning chains, and we evaluate on the last 1,000 questions from the development set. Bamboogle is a manually constructed benchmark that probes compositional reasoning, featuring two-hop questions whose supporting facts are present in Wikipedia but are challenging to retrieve together. We evaluate on the entire 125-question test set. MedQA comprises professional medical examination questions, such as those in the USMLE format, with four-way multiple-choice answers. We focus on the English split and evaluate on the standard 1,273-question test set. For process supervision, we sample 1,000 questions from the MedQA training set to construct process-reward trajectories, mirroring the approach used for HotpotQA.

For HotpotQA, 2WikiMultihopQA, and Bamboogle, we report Exact Match (EM) and token-level F1. For MedQA, we report accuracy.

\subsection{Retrieval corpora and information-retrieval environment}

For HotpotQA, 2WikiMultihopQA, and Bamboogle, agents retrieve evidence from English Wikipedia. For MedQA, we use a domain-specific corpus consisting of medical textbooks and StatPearls articles, pre-processed into a retrieval collection as described in prior work \cite{xiong2024benchmarking}

All agents interact with a unified hybrid retrieval pipeline. We combine lexical and dense retrieval using Reciprocal Rank Fusion \cite{cormack2009reciprocal}. For Wikipedia-based tasks, BM25 \cite{robertson2009probabilistic} serves as the lexical retriever and BGE-Base \cite{xiao2023cpack} as the dense retriever. On MedQA, we use BM25 together with MedCPT \cite{jin2023medcpt}. For each query, the top 32 documents from the fused ranking are retrieved, appended to the agent's history, and made available at subsequent reasoning steps

Retrieval hyperparameters, including retriever types, fusion scheme, and number of documents per query, remain fixed across agents. This ensures that observed performance differences reflect agent behavior rather than changes in the underlying information retrieval system

\subsection{RAG-Gym formulation and agent architectures}

\subsubsection{High-level Markov decision process formulation}

We frame agentic search as a high-level Markov decision process (MDP) where the agent interacts with the retrieval environment using discrete macro-actions

At each step $t$, the state is defined as
\begin{equation}
s_t = (Q, H_t),
\end{equation}
where $Q$ is the original question and $H_t$ is the information-seeking history
\begin{equation}
H_t = \{(q_1, D_1), \dots, (q_{t-1}, D_{t-1})\}.
\end{equation}
Each $q_i$ represents a search query and $D_i$ is the corresponding set of retrieved documents. The initial history is $H_1 = \varnothing$

The action space consists of both query actions and answer actions
\begin{equation}
A = A_q \cup A_p.
\end{equation}
Actions in $A_q$ correspond to issuing a search query, while actions in $A_p$ correspond to producing a candidate final answer. When the agent selects a query action $a_t \in A_q$, it is interpreted as a query $q_t$. The IR function $\mathrm{IR}(q_t)$ retrieves documents $D_t$, and the pair $(q_t, D_t)$ is appended to the history to form $s_{t+1}$. If the agent selects an answer action $a_t \in A_p$, the episode ends

Outcome rewards are assigned only to final answers. For a state-action pair $(s_t, a_t)$, the reward is defined as
\begin{equation}
R(s_t, a_t) =
\begin{cases}
0 & \text{if } a_t \in A_q \\
F(a_t, g(Q)) & \text{if } a_t \in A_p
\end{cases}.
\end{equation}
Here, $g(Q)$ denotes the ground-truth answer and $F$ is the dataset-specific evaluation metric such as EM, F1, or accuracy. The agent aims to maximize the expected discounted return over a trajectory. In practice, we focus on maximizing the average terminal reward across questions

This MDP formulation makes intermediate actions and reasoning steps explicit, supporting process-level supervision and critic-guided evaluation in addition to standard outcome-based rewards

\subsubsection{Functional decomposition}

We decompose agent capabilities into six functional modules. The first is answer generation, responsible for producing the final response. The second is question reasoning, which breaks down the problem into intermediate steps. The third is retrieval augmentation, which integrates retrieved content into the reasoning process. The fourth is query generation, which formulates effective search queries. The fifth is document summarization, which condenses retrieved passages into concise summaries. The sixth is reasoning reflection, which examines the current reasoning for unsupported claims and proposes targeted follow-up queries.

This modular decomposition establishes a unified framework for comparing agent architectures and attributing performance improvements to specific capabilities.

\subsubsection{Baseline agents}

All baseline agents share the same retrieval environment and backbone unless otherwise specified.

\paragraph{Direct.} The Direct agent prompts the language model to produce an answer in a single step. It does not perform explicit intermediate reasoning or retrieval and instantiates only the answer-generation module.

\paragraph{CoT \cite{wei2022chain}.} The CoT agent generates a step-by-step reasoning chain followed by an answer, all within a single iteration. It includes question reasoning but does not perform iterative retrieval or reflection.

\paragraph{RAG \cite{lewis2020retrieval}.} The RAG agent issues the original question as a search query and receives retrieved documents. It then generates an answer conditioned on this augmented context, resulting in a two-step pipeline with a single retrieval call.

\paragraph{ReAct \cite{yao2023react}.} ReAct alternates between natural-language reasoning and actions, which can be either search or answer. At each step, the model decides whether to continue searching or to answer, enabling multi-step interactive information seeking.

\paragraph{Search-o1 \cite{li2025search}.} Search-o1 extends ReAct by introducing a summarization stage. Retrieved documents are first condensed into short query-specific answers, and subsequent reasoning operates over these summaries instead of the raw documents.

\subsubsection{Re$^2$Search and Re$^2$Search++}

Our proposed Re$^{2}$Search agent instantiates all six functional modules, with reasoning reflection as the central addition. At each step, the reasoning model generates a step-by-step explanation and a tentative answer. It then identifies the first claim in the reasoning that is not supported by the current history and formulates this claim as a targeted query. This approach ensures that each search action addresses a specific knowledge gap in the answer construction process and reduces reliance on generic or heuristic queries.

Re$^2$Search++ refers to the complete system that integrates the Re$^2$Search architecture with process-level direct preference optimization (see Fig. \ref{fig:supervision_algorithm}a) and critic-guided inference (see Fig. \ref{fig:critic}a).

\subsection{Process-supervision data}

\subsubsection{Trajectory generation and filtering}

To obtain process-level supervision, we construct a dataset of state-action preferences from trajectories generated by untuned agents. For each question, we roll out multiple trajectories by sampling actions at each step from the current policy. We retain only those trajectories whose final answer matches the ground truth. This ensures that process supervision is applied along successful solution paths.

At each visited state, we sample multiple candidate next actions, which may be queries or answers. This collection of candidates forms the basis for preference annotation.

\subsubsection{Preference annotation and reward tuples}

Candidate actions at a given state are ranked, rather than scored independently, by an external annotator according to three criteria:
\begin{itemize}
\item {Sufficiency:} if the existing history already supports answering the question, an answer action is preferred over additional searches.
\item {Utility:} queries should be precise, actionable and directly relevant to resolving the question, avoiding unnecessary or tangential details.
\item {Non-redundancy:} actions that introduce new, informative content are preferred over those that repeat or slightly rephrase previous queries.
\end{itemize}

We primarily use GPT-4o as the annotator to produce ranked lists of candidate actions. For MedQA, a subset of trajectories is also annotated by medically trained co-authors. This enables us to measure agreement between human experts and reward models trained on GPT-4o preferences, as reported in Fig. \ref{fig:discussion}c.

The final process-reward dataset consists of tuples $(s, a^{+}, a^{-})$, where $a^{+}$ and $a^{-}$ denote the preferred and less-preferred actions for state $s$. This representation supports both direct policy optimization and the training of separate reward models.

\subsection{Post-training objectives for the actor}

We investigate three post-training objectives defined over the process-reward dataset: supervised fine-tuning (SFT), direct preference optimization (DPO), and proximal policy optimization (PPO). Let $D$ denote the set of state-action pairs or triples derived from the preference data

\paragraph{Supervised fine-tuning (SFT) \cite{ouyang2022training}.}
SFT treats preferred actions as demonstrations and maximizes their log-likelihood under the policy:
\begin{equation}
\mathcal{L}_{\text{SFT}}(\theta)
= - \mathbb{E}_{(s, a^{+}) \sim D}
\left[ \log \pi_{\theta}(a^{+} \mid s) \right].
\end{equation}
This objective encourages the actor to imitate high-quality trajectories but does not explicitly model less-preferred alternatives

\paragraph{Direct preference optimization (DPO) \cite{rafailov2023direct}.}
DPO reformulates each preference as a pair $(a^{+}, a^{-})$ for a shared state $s$ and encourages the policy to favor $a^{+}$ over $a^{-}$ relative to a reference policy $\pi_{\text{ref}}$:
\begin{equation}
\mathcal{L}_{\text{DPO}}(\theta)
= - \mathbb{E}_{(s, a^{+}, a^{-}) \sim D}
\left[
\log \sigma\!\left(
\beta \Delta_{\theta}(s, a^{+}, a^{-})
\right)
\right],
\end{equation}
where
\begin{equation}
\Delta_{\theta}(s, a^{+}, a^{-})
= \log \frac{\pi_{\theta}(a^{+} \mid s)}{\pi_{\text{ref}}(a^{+} \mid s)}
- \log \frac{\pi_{\theta}(a^{-} \mid s)}{\pi_{\text{ref}}(a^{-} \mid s)}
\end{equation}
Here, $\beta$ is a temperature parameter and $\sigma$ is the sigmoid function

\paragraph{Proximal policy optimization (PPO) \cite{schulman2017proximal}.}
For PPO, we first train a process reward model $r_{\phi}(s,a)$ on the preference data (Section~\ref{sec:critic}). The actor is then updated by maximizing the expected process reward of newly generated actions with a clipped policy-gradient objective, using $r_{\phi}$ as the reward signal.

For SFT and DPO, we apply Low-Rank Adaptation (LoRA) \cite{hu2021lora} to the backbone with rank $r=256$ and scaling factor $\alpha=512$ on all attention layers, implemented using the TRL library \cite{vonwerra2022trl}. PPO is implemented with OpenRLHF \cite{hu2024openrlhf} and full-parameter tuning of the actor. For Search-o1 and Re$^{2}$Search, only the reasoning language model is tuned, while the knowledge-summarization model remains frozen. Detailed optimization hyperparameters are provided in the code release.

\subsection{Critic training and critic-guided inference}

We train an external critic to predict process-level preferences between candidate actions. The critic shares the same backbone family as the actor and is trained with a Bradley-Terry-style pairwise objective \cite{bradley1952rank}. Given a state $s$ and two actions $a^{+}$ and $a^{-}$, the critic is optimized to assign a higher score to $a^{+}$ by maximizing
\begin{equation}
\mathcal{J}(\phi) = \mathbb{E}_{(s, a^{+}, a^{-}) \sim D} \left[ \log \sigma\!\big(r_{\phi}(s, a^{+}) - r_{\phi}(s, a^{-})\big) \right].
\end{equation}
Here, $\sigma$ denotes the sigmoid function, and $r_{\phi}(s, a)$ represents the scalar score predicted by the critic for action $a$ in state $s$.

At inference time, when critic guidance is enabled, we use a Best-of-$N$ strategy. For each state $s_t$, the actor samples $N$ candidate actions $\{a_1, \dots, a_N\}$. The critic scores each candidate, and the action with the highest score is executed. If the selected action is a query, the environment returns retrieved documents and the process continues. If it is a final answer, the episode terminates

Unless otherwise specified, we use $N=10$ samples per step. We set the temperature to 1.0 for critic-guided runs and to 0.0 when no critic is used. Algorithm \ref{alg:prm_best_of_n} summarizes the critic-guided inference procedure.

\begin{algorithm}\small
\caption{Critic-Guided Inference (Best-of-$N$)}
\label{alg:prm_best_of_n}
\begin{enumerate}
    \item \textbf{Input:} Question $Q$, actor $\pi_{\theta}$, critic $r_{\phi}$, number of actions $N$, maximum steps $T$, IR function \textit{IR}.
    \item \textbf{Initialize} state $S \leftarrow (Q, H_1 = \emptyset)$.
    \item \textbf{For} $t = 1$ to $T$:
    \begin{enumerate}
        \item Generate $N$ actions: $a_1,\cdots,a_N \sim \pi_{f(\theta)}(\cdot|S)$.
        \item Compute process rewards and select the best action: $a^* \leftarrow \arg\max_{a \in \{a_1,\cdots,a_N\}} r_{\phi}(S, a) $.
        
        \item \textbf{If} $a^*$ is a search query:
        \begin{enumerate}
            \item Retrieve documents: $D \leftarrow \textit{IR}(a^*)$.
            \item Update state: $S \leftarrow (Q, H_{t+1} = H_{t} \cup \{(a^*, D)\})$.
        \end{enumerate}
        
        \item \textbf{If} $a^*$ is a final answer:
        \begin{enumerate}
            \item Return $a^*$ and terminate the process.
        \end{enumerate}
    \end{enumerate}
    \item \textbf{End For}
\end{enumerate}
\end{algorithm}

\subsection{Backbone models, prompts and implementation details}

Unless otherwise specified, both actor and critic use Llama-3.1-8B-Instruct \cite{dubey2024llama} as the backbone. To study cross-model transfer, we additionally evaluate settings where a critic trained on Llama-3.1-8B trajectories guides the inference of a GPT-4o-mini \cite{hurst2024gpt} actor without retraining the critic. We also implemented Re$^2$Search++ with Qwen-2.5-7B-Instruct for a fair comparison with outcome-supervised agents (Fig. \ref{fig:prm_vs_orm}).

Re$^2$Search relies on the structured prompt templates shown below for history summarization and action generation. The history-summarization prompt in Prompt \ref{fig:prompt_summarize} enforces concise, factual summaries of retrieved documents that can be reused in subsequent queries. The action-generation prompt for Re$^2$Search in Prompt \ref{fig:prompt_research} couples step-by-step answer construction with the explicit identification of unverified claims, which are then translated into targeted follow-up queries. Prompt \ref{fig:prompt_rank} is used with GPT-4o to obtain process-level preference annotations for training the process reward models.
\begin{AIbox}{Prompt template for history knowledge summarization in Search-o1 and Re$^2$Search}
\label{fig:prompt_summarize}
You are a helpful assistant tasked with answering a follow-up query using the relevant documents provided.\\

\#\#\# Relevant Documents\\
\verb|{{documents}}|\\

\#\#\# Context\\
Original question: \verb|{{question}}|\\

\#\#\# Follow-up Query\\
\verb|{{query}}|\\

Answer the follow-up query succinctly, using only the information from the documents. When the documents do not provide sufficient information, explicitly point this out instead of making up facts. Do not include unrelated or excessive details in the response.
\end{AIbox}
\begin{AIbox}{Prompt template for generating actions using the Re$^2$Search agent}
\label{fig:prompt_research}
You are a helpful assistant. Your task is to answer a given question following user instructions.'\\

\#\#\# Information-seeking History\\
\verb|{{history}}|\\

\#\#\# Original Question\\
\verb|{{question}}|\\

Your output must include three sections:\\
1. **\#\#\# Step-by-step Reasoning**:\\
  - Think step-by-step and then answer the question.\\

2. **\#\#\# Unverified Claim Identification**:\\
  - Identify if there are claims in the step-by-step reasoning section that are not grounded in the information-seeking history section.\\
  - If yes, summarize the first piece of missing information as an atomic query to search in an external knowledge base.\\
  - If no, clearly state that no further query is needed.\\

3. **\#\#\# Structured Output**:\\
  - Present your predicted answer and generated query (if applicable) in the following JSON format:\\
    ```json\\
    \{\\
        ``predicted\_answer": ``Provide a single letter (for multiple-choice questions), digit, word, or short phrase here.",\\
        ``generated\_query": ``Provide an entity, question, or statement to be searched in an external knowledge base. Output \textbackslash``None\textbackslash" if no query is generated.",\\
    \}\\
    ```
\end{AIbox}
\begin{AIbox}{Prompt template for ranking candidate actions with GPT-4o}
\label{fig:prompt_rank}
You are a decision-evaluation assistant. Your task is to rank the proposed actions from the most appropriate to the least appropriate as the next step in a sequential decision-making process aimed at solving a given question.\\

\#\#\# Original Question:\\
\verb|{{question}}|\\

\#\#\# Information-Seeking History:\\
\verb|{{curr_history}}|\\

\#\#\# Proposed Next Actions:\\
\verb|{{actions_text}}|\\

\#\#\# Important Assumption\\
The agent has no prior knowledge about the subject matter. It must rely solely on the information-seeking history provided to evaluate and answer the original question. Assumptions not explicitly supported by the history must not influence the ranking of proposed actions.\\

\#\#\# Evaluation Criteria for Appropriateness\\
1. **Sufficiency Check**:\\
- Determine whether the available information is sufficient to directly answer the original question. If not, the proposed action to ``Answer'' is inappropriate.\\
- Prioritize queries that gather specific, missing information essential to solving the question.\\
- If the history already contains all necessary information, then ``Answer'' is the most appropriate action, and the correct answer should be ranked highest.\\

2. **Utility Check**:\\
- Queries must be precise, actionable, and directly relevant to solving the question.\\
- Prioritize foundational queries that establish critical context or general knowledge necessary for more specific follow-ups.\\
- Rank overly narrow or prematurely specific queries lower if they presume knowledge not yet available.\\
- Avoid irrelevant queries that do not contribute to solving the original question.\\

3. **Redundancy Check**:\\
- Queries that duplicate information already covered in the history or repeat previous queries should be ranked lower.\\
- Proposed actions must add new value to the decision-making process by seeking new or clarifying missing information.\\

\#\#\# Expected Output Format\\
- Output the indices of the ranked actions in JSON format: ```json\{``ranked\_indices'': [list of indices]\}'''.\\
- Rank actions from most appropriate to least appropriate based on the evaluation criteria above.\\
- Do not provide additional explanations or reasoning.'''
\end{AIbox}

\subsection{Data and Code availability}

The HotpotQA, 2WikiMultihopQA, Bamboogle, and MedQA datasets used in this study are publicly available from their original sources. Retrieval corpora are Wikipedia (for HotpotQA/2Wiki/Bamboogle) and the MedRAG preprocessed collection comprising medical textbooks and StatPearls (for MedQA) used in previous work \cite{xiong2024benchmarking}. Trained agent and critic models are available at \url{https://huggingface.co/RAG-Gym}.
Code for RAG-Gym implementations is available at \url{https://github.com/RAG-Gym/RAG-Gym}.

\bibliography{reference}

\section{Acknowledgments}
This research was partially supported by the Intramural Research Program of the National Institutes of Health (NIH). The contributions of the NIH author(s) are considered Works of the United States Government. This research was also partially supported by the NIH Pathway to Independence Award K99LM014903 (Q.J.), NIH R01LM014012, and NSF IIS-2106913. The findings and conclusions presented in this paper are those of the author(s) and do not necessarily reflect the views of the NSF, the NIH, or the U.S. Department of Health and Human Services.

\end{document}